\pdfoutput=1
\documentclass[11pt]{article}

\usepackage{acl}

\usepackage{times}
\usepackage{latexsym}
\usepackage{float}
\usepackage{booktabs}
\usepackage{graphicx}
\graphicspath{ {./images/} }
\usepackage{xcolor}

\usepackage{siunitx} \usepackage{float}

\definecolor{contextOrange}{HTML}{ffab40}
\definecolor{bodyBlue}{HTML}{4285f4}
\definecolor{returnBlue}{HTML}{0097a7}
\definecolor{outputRed}{HTML}{FF0000}
\usepackage{float}
\usepackage{multirow}
\usepackage{colortbl}
\usepackage{pifont}
\usepackage[T1]{fontenc}
\usepackage[utf8]{inputenc}

\usepackage{microtype}

\title{Evaluating How Fine-tuning on Bimodal Data Effects Code Generation}

\author{Gabriel Orlanski \\
  \texttt{go533@nyu.edu} \\ \And
  Seonhye Yang \\
  \texttt{sy3420@nyu.edu} \\ \And
  Michael Healy \\
  \texttt{mbh425@nyu.edu}
  }

\begin{document}
\maketitle
\begin{abstract}
Despite the increase in popularity of language\-models for code generation, it is still unknown how training on bimodal coding forums affects a model's code generation performance and reliability. We, therefore, collect a dataset of over 2.2M StackOverflow questions with answers for fine\-tuning. These fine\-tuned models have average $pass@k$ improvements of 54.64\% and 85.35\% on the HumanEval \citep{chen_evaluating_2021} and Mostly Basic Program Problems \citep{austin_program_2021} tasks, respectively. This regime further decreases the number of generated programs with both syntax and runtime errors. However, we find that at higher temperatures, there are significant decreases to the model's ability to generate runnable programs despite higher $pass@k$ scores, underscoring the need for better methods of incorporating such data that mitigate these side effects. The code can be found \href{https://github.com/gabeorlanski/bimodal-code-generation}{https://github.com/gabeorlanski/bimodal-code-generation}
\end{abstract}

\section{Introduction}
\begin{figure}[t]
    \centering
    \includegraphics[width=1\linewidth,keepaspectratio]{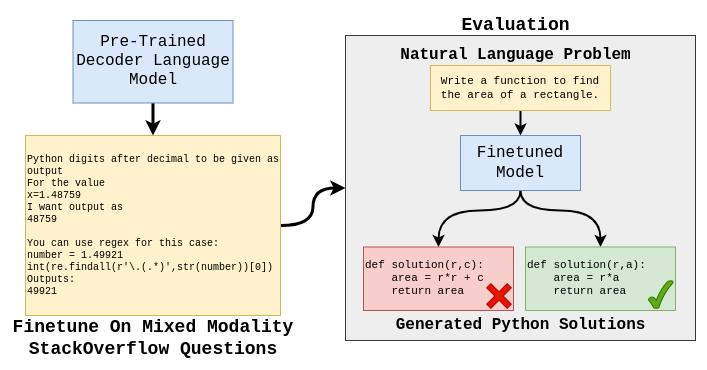}
    \caption{Overview of our approach paper. We first take a pretrained decoder only language model checkpoint and fine-tune it on mixed modality question and answers from StackOverflow. We then take this model and evaluate its code generation performance on HumanEval and Mostly Basic Program Problems (MBPP) datasets. For each, we generate 200 programs per problem and check their functional correctness with a set of test cases to calculate their $pass@k$ scores.}
    \label{fig:overview}
\end{figure}
The past few years have seen a rapid increase in interest in using pretrained language models(LM) for program synthesis. Recent works from both \citet{austin_program_2021} and \citet{chen_evaluating_2021} evaluated how large transformer models \citep{vaswani2017attention} perform on synthesizing Python programs from a natural language description. Subsequent work has further evaluated how these models perform in coding competitions \citep{li_competition-level_2022} and how pretraining on a diverse set of programming languages affect downstream performance \citep{xu_systematic_2022}. While much focus has been on how to use GitHub\footnote{https://github.com/} data, educational resources such as Stack Overflow (SO)\footnote{https://stackoverflow.com/}, where natural language and programming language occur together in a significantly different setting than the documentation that occurs in source code, have been under-explored.

Some works have used SO as training data for generation tasks \citep{yin_learning_2018,gao2020pile,Orlanski2021ReadingSE}, yet it is not until recently that there has been interest in utilizing this data for downstream tasks that require code execution \citep{friedIncoder}. However, the effects of training on a bimodal corpus with both natural language (NL) and programming language (PL) are still unknown. One potential reason is that there are no guarantees that the code found has valid syntax, unlike the source code. Such concerns have inspired works that aim to improve the reliability of code generation with transformers \citep{guo2022learning,poesia2022synchromesh}. Such solutions, unfortunately, require a significant amount of engineering. In this paper, we, therefore, aim to answer two questions: \textit{(1) Does fine-tuning a pretrained decoder language model on unstructured bimodal SO data improve its text-to-code generation performance? (2) How does fine-tuning on this data impact the model's ability to generate valid code?}

To this end, we first collect and parse over 2.2M python questions with their answers into a dataset comprising over 1.17B tokens. As our focus is on generating Python programs, we do not examine how bimodal data with different programming languages affect downstream performance. We then fine-tune both CodeParrot 110M \citep{Thoppilan2022LaMDALM} and GPT Neo \citep{gpt-neo} models on this data. We then evaluate these models on the HumanEval and MBPP datasets. We find average $pass@k$ improvements of 52.40\% and 87.59\% for CodeParrot and GPT Neo, respectively. 

Next, we perform ablations in which we remove a modality prior to fine-tuning. We find that the models fine-tuned on both modalities have, on average, 20.33\% better $pass@k$ scores than those fine-tuned without answer code. Further, the bimodal models are only 6.11\% worse than those trained without natural language, suggesting that these models can learn without any indication of modality, but it is not an optimal environment. We conclude with an error analysis of the generated programs and find that fine-tuning this data increases the number of errors when evaluating in a zero-shot setting.

\section{Related Work}
% \subsection{Natural Language to Code}

Early work on translating NL to PL focused on either generating database queries \citep{Zelle1996LearningTP,Iyer2019LearningPI} or domain specific languages \citep{dong2016language}. Recent years have seen an increased interest in generating open-ended languages such as Java \citep{iyer_mapping_2018} and Python \citep{clement_pymt5_2020, roziere_dobf_2021}. \citet{yin2017syntactic} proposed a neural model in combination with a transition system to generate abstract syntax trees \citep{yin_tranx_2018}. With the advent of the transformer \citep{vaswani2017attention}, the focus has shifted from using code's inherent structure to treating it as raw text \citep{wang2021codet5,ahmad_unified_2021,clement_pymt5_2020}. \citet{feng2020codebert} proposed using a masked language model with bimodal text from the CodeSearchNet challenge \citep{husain2019codesearchnet}. \citet{austin_program_2021} and \citet{chen_evaluating_2021} both propose using a decoder language model \citep{brown2020language} trained on large amounts of source and web code. Finally, \citet{xu_systematic_2022} perform a systematic evaluation of these models, finding that having some natural language in the training corpus helps with general code-language modeling.

\begin{table*}[ht]
    \centering
 \begin{tabular}{l|ccc|ccc}
\toprule
 &   \multicolumn{3}{c}{HumanEval} & \multicolumn{3}{c}{MBPP} \\
Model   & @1 & @10 & @100 & @1 & @10 & @80 \\
\midrule
Codex 300M$^a$& 13.17 & 20.37 & 36.27 & --- &---  &  --- \\
PolyCoder 2.7B$^b$ & 5.60 & 9.80 & 27.70 & --- &---& ---\\
InCoder 6.7B$^c$ &  15.20 & 27.80 & 47.00 & 19.40&--- & --- \\
Davinci Codex$^a$ $^*$ & 36.99 & --- & 81.70 & 50.40 &---  &  84.40 \\
PaLM Coder 540B$^d$ &36.00 & --- & 88.40 & 47.00 & --- & 80.80\\
\midrule
CodeParrot 110M & 3.80 & 6.40 & 10.84 & 2.50 & 11.55 & 27.58 \\
\ \ +SO & \bfseries \bfseries \bfseries \bfseries 5.53 & 8.83 & 15.78 & 5.80 & 19.18 & 37.15 \\
\ \ +SO -Code & 3.54 & 7.51 & 14.11 & 4.83 & 15.73 & 32.48 \\
\ \ +SO -NL & 5.45 & \bfseries \bfseries \bfseries \bfseries 10.02 & \bfseries \bfseries \bfseries \bfseries 17.54 & \bfseries \bfseries \bfseries \bfseries 6.09 & \bfseries \bfseries \bfseries \bfseries 20.06 & \bfseries \bfseries \bfseries \bfseries 38.06 \\
 \midrule
GPT Neo 125M & 0.83 & 1.51 & 3.43 & 0.33 & 2.32 & 8.90 \\
\ \ +SO & 1.49 & 2.82 & \bfseries \bfseries \bfseries \bfseries 5.26 & 0.82 & 5.45 & 16.58 \\
\ \ +SO -Code & 0.73 & 2.36 & 4.57 & 0.97 & 4.63 & 14.42 \\
\ \ +SO -NL & \bfseries \bfseries \bfseries \bfseries 1.69 & \bfseries \bfseries \bfseries \bfseries 3.07 & 5.04 & \bfseries \bfseries \bfseries \bfseries 1.11 & \bfseries \bfseries \bfseries \bfseries 5.92 & \bfseries \bfseries \bfseries \bfseries 17.36 \\
\bottomrule
\end{tabular}

    \caption{$pass@k$ Results on both the HumanEval and MBPP task. Best reported results from three runs with $T \in \{0.2,0.6,0.8\}$, and $p=0.95$ and taking the best values for each $k$. We additionally include results reported by prior works. The \textbf{bolded} entries are the best value for their respective column and model in the \textit{Model} column. $^a$\citet{chen_evaluating_2021} $^b$\citet{xu_systematic_2022} $^c$\citet{friedIncoder} $^c$\citet{chen_evaluating_2021} $^d$\citet{chowdhery2022palm} $^*$ results calculated by \citet{chowdhery2022palm} following the same settings outlined in \citet{chen_evaluating_2021}. \autoref{tab:full_results} has the full results for more $k$ values.}
    \label{tab:baselines}
\end{table*}
\section{Method}\label{sec:so}
% \subsection{Stack Overflow Data Collection}
To utilize the SO data dump\footnote{https://archive.org/details/stackexchange} efficiently, we first align the questions with their given answers while filtering out questions that do not have a Python-related tag. This results in more than 2.2M questions for nearly 7GB of data. Including the answers from each question, this translates to approximately 1.17B tokens used for fine-tuning. For every question, we sort the answers using the following strategy: if an accepted answer exists, put it first and sort the remaining answers by their user score. We do not examine other sorting strategies or dataset filtering based on attributes besides the tags as it is outside this paper's scope. 

As shown in as shown in \autoref{fig:overview}, we remove all HTML tags from the SO text. Prior work using SO as a training corpus has explicitly removed code \citep{ahmad_unified_2021}. We choose not to remove the code as there is evidence that code is the \textit{most} important modality contained in SO questions \citep{Orlanski2021ReadingSE}. This line of work is similar to \citet{austin_program_2021} in that it both focuses on examining the effects of training LMs on code that is \textit{not} from source code. However, they do \textit{not} remove HTML tags from their training corpus and use more resources than only SO. Our focus differs in that we limit the scope to only the case where the training text does not contain modality markers. We finally pack the title, question body, and answer text into a single sequence for training.

% Given that we are using a decoder transformer model, we combine the title, question body, and answers sorted by their score into a single sequence for causal language modeling. Formally, we combine the question $Q$ with body $Q_B$ and title $Q_T$, with answers $A = [a_1,\ldots,a_n]$ into the  sequence $Q_T+Q_B+a_1+\ldots+a_n$. We finally append an \verb|<EOS>| token to each sequence.   

\section{Experimental Setup}\label{sec:experiment}
We fine-tune the CodeParrot \citep{tunstall2022natural} and GPT-Neo \citep{gpt-neo} models with parameter counts 110M and 125M respectively on the SO data using 4 RTX8000 GPUs with a batch size of 32 and a sequence length of 1024. We selected these two models as they are both decoder LMs with very different pretraining data regimes. CodeParrot was only trained on GitHub, whereas GPT-Neo was trained on the PILE \citep{gao2020pile} which includes a diverse range of sources. We trained for 25K steps with a starting learning rate of \num{5e-5} and a cosine scheduler with 750 warmup steps. We use the AdamW optimizer \citep{loshchilov2017decoupled} and the Transformers library \citep{wolf-etal-2020-transformers}.

We then evaluate these models on HumanEval \citep{chen_evaluating_2021} and MBPP \citep{holtzman2019curious} by generating 200 programs for each problem in both datasets. Performance is then measured by checking if the programs pass given test cases, the results of which are then used to calculate $pass@k$ \citep{chen_evaluating_2021}. Following prior works, we use $k\in \{1,10,100\}$ for HumanEval \citep{chen_evaluating_2021} and $k \in \{1,10,80\}$ for MBPP \citep{chowdhery2022palm, austin_program_2021}.\footnote{In \autoref{tab:full_results} and for the averages we use $k\in\{1,10,50,80,100\}$. } Unless $k$ is explicitly stated, we report improvements in $pass@k$ across ablations as the mean percent change of the $pass@k$ scores for $k\in\{1,10,50,80,100\}$. Additionally, for MBPP, we follow \citep{austin_program_2021} by fine-tuning the models on the training split for 100 steps. For this, we use a batch size of 32, a context window of 1024 tokens, and a learning rate of \num{1e-5}. We prompt the model with the problem in the task-specific formatting. For HumanEval, this is the function signature and the docstring, while for MBPP, it is the NL problem with the three test cases. We follow \citet{austin_program_2021} in providing the model an NL prompt in addition to the problem statement. As it is not the focus of this paper, we do not examine how changing this prompt affects the performance. 

\section{Results}\label{sec:results}
\autoref{tab:baselines} shows our results as well as those reported by recent work on the HumanEval and MBPP datasets. We find that fine-tuning CodeParrot on SO data yields average $pass@k$ improvements of 44.95\% and 60.85\% compared to the baseline model results on HumanEval and MBPP, respectively. Fine-tuning GPT Neo on SO data yields a more significant improvement of 65.34\% and 109.85\% on HumanEval and MBPP, respectively.

\begin{table}[h]
    \centering
\begin{tabular}{l|ccc|c}
\toprule
 Model& $S$ & $R$ & $T$ & $C$ \\
\midrule
CodeParrot & {\cellcolor[HTML]{6AAE4D}} \color[HTML]{000000} 22.7 & {\cellcolor[HTML]{EC7D82}} \color[HTML]{000000} 48.0 & {\cellcolor[HTML]{EC7D82}} \color[HTML]{000000} 122.5 & {\cellcolor[HTML]{F0C1C3}} \color[HTML]{000000} 6.0 \\
+SO & {\cellcolor[HTML]{EC7D82}} \color[HTML]{000000} 43.5 & {\cellcolor[HTML]{6AAE4D}} \color[HTML]{000000} 42.9 & {\cellcolor[HTML]{6AAE4D}} \color[HTML]{000000} 105.4 & {\cellcolor[HTML]{7DB864}} \color[HTML]{000000} 7.5 \\
\ \ -Code & {\cellcolor[HTML]{EE969A}} \color[HTML]{000000} 41.6 & {\cellcolor[HTML]{F2E6E7}} \color[HTML]{000000} 46.2 & {\cellcolor[HTML]{76B45B}} \color[HTML]{000000} 106.0 & {\cellcolor[HTML]{EC7D82}} \color[HTML]{000000} 5.5 \\
\ \ -NL & {\cellcolor[HTML]{F2F2F2}} \color[HTML]{000000} 32.7 & {\cellcolor[HTML]{F2E5E5}} \color[HTML]{000000} 46.2 & {\cellcolor[HTML]{F2F2F2}} \color[HTML]{000000} 112.6 & {\cellcolor[HTML]{6AAE4D}} \color[HTML]{000000} 7.7 \\
\midrule
GPT Neo & {\cellcolor[HTML]{EC7D82}} \color[HTML]{000000} 65.8 & {\cellcolor[HTML]{EC7D82}} \color[HTML]{000000} 86.4 & {\cellcolor[HTML]{6AAE4D}} \color[HTML]{000000} 46.6 & {\cellcolor[HTML]{EC7D82}} \color[HTML]{000000} 0.9 \\
+SO & {\cellcolor[HTML]{C3DDB8}} \color[HTML]{000000} 37.9 & {\cellcolor[HTML]{8FC27A}} \color[HTML]{000000} 75.6 & {\cellcolor[HTML]{ED8D92}} \color[HTML]{000000} 84.3 & {\cellcolor[HTML]{A3CC91}} \color[HTML]{000000} 1.8 \\
\ \ -Code & {\cellcolor[HTML]{F2F2F2}} \color[HTML]{000000} 44.1 & {\cellcolor[HTML]{6AAE4D}} \color[HTML]{000000} 74.3 & {\cellcolor[HTML]{EFACAF}} \color[HTML]{000000} 80.0 & {\cellcolor[HTML]{F0CED0}} \color[HTML]{000000} 1.2 \\
\ \ -NL & {\cellcolor[HTML]{6AAE4D}} \color[HTML]{000000} 27.5 & {\cellcolor[HTML]{F0C4C6}} \color[HTML]{000000} 83.4 & {\cellcolor[HTML]{EC7D82}} \color[HTML]{000000} 86.7 & {\cellcolor[HTML]{6AAE4D}} \color[HTML]{000000} 2.0 \\
\bottomrule
\end{tabular}

    \caption{Average number of programs generated per problem in HumanEval that have $S$yntax/$R$untime errors, failed the $T$ests, or $C$orrectly solved the problem. Mean was taken from all programs generated with temperature in $\{0.2,0.6,0.8\}$. {\colorbox[HTML]{61A046}{\color[HTML]{F1F1F1} Green}} values are better than {\colorbox[HTML]{EB666D}{\color[HTML]{F1F1F1} Red}} ones.}
    \label{tab:errors}
\end{table}
\subsection{Removing Modalities}
Given that there are multiple modalities present in the SO data, we aim to quantify the impact that each has on an LM's program synthesis ability. To this end, we evaluate the $pass@k$ performance after fine-tuning the pre-trained models on the SO data with a specific modality removed. We use the same experimental setup described in \autoref{sec:experiment} but remove either the Code or the NL from the answers. Across all runs, we do not modify either the title or the question body. 

The CodeParrot fine-tuned on SO outperforms the same checkpoint fine-tuned on SO, with the answer code removed by an average of 21.88\% and 17.16\% on HumanEval and MBPP, respectively. However, fine-tuning on both modalities performs worse than the model fine-tuned without answer NL by an average of 7.57\% on HumanEval and 3.42\% on MBPP. The same is true for GPT Neo, as fine-tuning on both modalities is 3.74\% and 3.42\%, worse than when only the answer code is used on HumanEval and MBPP, respectively. This indicates that while LMs can learn from unstructured bimodal data, the NL in SO acts as noise. 
\begin{figure}[h]
    \centering
    \includegraphics[width=1\linewidth,keepaspectratio]{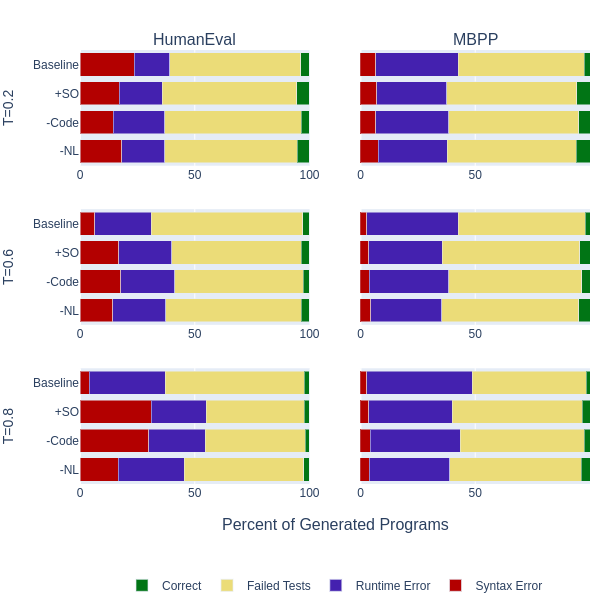}
    \caption{Breakdown of execution results for programs generated by CodeParrot across temperatures $\{0.2,0.6,0.8\}$. HumanEval has 164 problems and MBPP has 500 problems. As we generate 200 programs per problem, we have a total of 32,800 and 100,000 programs respectively. }
    \label{fig:shift}
\end{figure}
\subsection{Error Analysis}
As shown in \autoref{tab:errors}, fine-tuning both CodeParrot and GPT Neo on the full SO data results in a decrease in the number of programs with runtime errors generated per problem. A plausible explanation for this is that StackOverflow is primarily for developers to find help on programming problems and, thus, specific "triggers" are likely commonly asked about. However, as shown in \autoref{fig:shift}, the bimodal setting and general lack of structural guarantees cause an increase in program errors as the temperature increases. When $T=0.8$, 31.13\% of the program generated by the model fine-tuned on SO has invalid syntax. The cause is not the inclusion of the NL, as 16.66\% of the programs generated by the model fine-tuned on SO without NL have syntax errors. One potential remedy to the situation is to fine-tune the model further on a solely code dataset such as MBPP, as shown in the right-hand column of \autoref{fig:shift}.  
\section{Conclusion}
In this paper, we evaluate the program synthesis abilities of pretrained decoder language models that have been fine-tuned on bimodal StackOverflow data. Our results show that this fine-tuning regime does indeed improve $pass@k$ scores by an average of 70\%. We then perform ablations where we remove each modality, finding that models fine-tuned on the bimodal data outperforms those fine-tuned on only the NL by an average of 20.33\% but is 6.11\% worse than those fine-tuned on only the answer code. We conclude with an error analysis of the generated programs and find that the number of programs with runtime errors drop by 11.49\% in HumanEval but syntax errors rise as temperature increases. With this in mind, future work should focus on how to keep the benefits of fine-tuning on StackOverflow while mitigating the negative side effects.

% \section*{Collaboration Statement}

% \textbf{Gabe Orlanski} wrote the StackOverflow parsing, experiment, evaluation, and analysis code.  He also ran all of the experiments and worked on the paper writing.

% \vspace{.05in}

% \noindent\textbf{Michael Healy} worked on paper writing and researching relevant works.

% \vspace{.05in}

% \noindent\textbf{Seonhye Yang} parsed tutorial libraries and worked on some of the high level analysis. 

\section*{Acknowledgment}

We thank He He, Angelica Chen, Jacob Austin, and Xaiver Garcia for their comments and helpful discussions. 

% \section{Conclusion}

% \bibliographystyle{acl_natbib}
\bibliography{bibliography}

\appendix
\begin{table*}[ht]
    \centering
 \begin{tabular}{l|ccccc|ccccc}
\toprule
& \multicolumn{5}{c}{HumanEval} & \multicolumn{5}{c}{MBPP} \\
Model & @1 & @10 & @50 & @80 & @100 & @1 & @10 & @50 & @80 & @100 \\
\midrule
CodeParrot 110M & 3.80 & 6.40 & 9.35 & 10.38 & 10.84 & 2.50 & 11.55 & 23.71 & 27.58 & 29.42 \\
 +SO & \bfseries \bfseries \bfseries \bfseries 5.53 & 8.83 & 13.69 & 14.97 & 15.78 & 5.80 & 19.19 & 32.93 & 37.15 & 39.02 \\
\ \ -Code & 3.54 & 7.51 & 12.13 & 13.50 & 14.11 & 4.83 & 15.77 & 28.18 & 32.48 & 34.63 \\
\ \ -NL & 5.45 & \bfseries \bfseries \bfseries \bfseries 10.02 & \bfseries \bfseries \bfseries \bfseries 14.96 & \bfseries \bfseries \bfseries \bfseries 16.46 & \bfseries \bfseries \bfseries \bfseries 17.54 & \bfseries \bfseries \bfseries \bfseries 6.09 & \bfseries \bfseries \bfseries \bfseries 20.06 & \bfseries \bfseries \bfseries \bfseries 34.17 & \bfseries \bfseries \bfseries \bfseries 38.05 & \bfseries \bfseries \bfseries \bfseries 39.85 \\
\midrule
GPT Neo 125M & 0.83 & 1.51 & 2.71 & 3.18 & 3.43 & 0.33 & 2.32 & 6.85 & 8.90 & 9.96 \\
+SO & 1.49 & 2.82 & 4.20 & 4.82 & \bfseries \bfseries \bfseries \bfseries 5.26 & 0.82 & 5.45 & 13.53 & 16.58 & 18.10 \\
\ \ -Code & 0.73 & 2.36 & 3.70 & 4.25 & 4.57 & 0.97 & 4.63 & 11.80 & 14.42 & 15.71 \\
\ \ -NL & \bfseries \bfseries \bfseries \bfseries 1.69 & \bfseries \bfseries \bfseries \bfseries 3.07 & \bfseries \bfseries \bfseries \bfseries 4.34 & \bfseries \bfseries \bfseries \bfseries 4.84 & 5.04 & \bfseries \bfseries \bfseries \bfseries 1.11 & \bfseries \bfseries \bfseries \bfseries 5.92 & \bfseries \bfseries \bfseries \bfseries 14.31 & \bfseries \bfseries \bfseries \bfseries 17.36 & \bfseries \bfseries \bfseries \bfseries 18.93 \\
\bottomrule
\end{tabular}

    \caption{Full $pass@k$ Results on both the HumanEval and MBPP task. Best reported results from three runs with $T \in \{0.2,0.6,0.8\}$, and $p=0.95$ and taking the best values for each $k$. We additionally include results reported by prior works. The \textbf{bolded} entries are the best value for their respective column and model in the \textit{Model} column.}
    \label{tab:full_results}
\end{table*}

\end{document}